\documentclass{article}

\usepackage{microtype}
\usepackage{graphicx}
\usepackage{subfigure}
\usepackage{booktabs} % for professional tables

\usepackage{hyperref}

\usepackage[accepted]{tccmlicml2021}

\icmltitlerunning{Forecasting emissions through Kaya identity using Neural ODEs}

\begin{document}

\twocolumn[
\icmltitle{Forecasting emissions through Kaya identity \\using Neural Ordinary Differential Equations}




\begin{icmlauthorlist}
\icmlauthor{Pierre Browne}{ICL}
\icmlauthor{Aranildo R. Lima}{AquaticInformatics}
\icmlauthor{Rossella Arcucci}{ICL}
\icmlauthor{César Quilodrán-Casas}{ICL}
\end{icmlauthorlist}

\icmlaffiliation{ICL}{Data Science Institute, Imperial College London}
\icmlaffiliation{AquaticInformatics}{Aquatic Informatics, Vancouver, Canada}

\icmlcorrespondingauthor{Pierre Browne}{pierre.browne20@imperial.ac.uk}
% \icmlcorrespondingauthor{Aranildo R. Lima}{aranildo.lima@aquaticinformatics.com}
% \icmlcorrespondingauthor{Rossella Arcucci}{r.arcucci@imperial.ac.uk}
% \icmlcorrespondingauthor{César Quilodrán}{cesar.quilodran-casas13@imperial.ac.uk}

\icmlkeywords{Machine Learning, ICML}

\vskip 0.3in
]



\printAffiliationsAndNotice{\icmlEqualContribution} % otherwise use the standard text.

\begin{abstract}
Starting from the Kaya identity, we used a Neural ODE model to predict the evolution of several indicators related to carbon emissions, on a country-level: population, GDP per capita, energy intensity of GDP, carbon intensity of energy. We compared the model with a baseline statistical model - VAR - and obtained good performances. We conclude that this machine-learning approach can be used to produce a wide range of results and give relevant insight to policymakers.
\end{abstract}

\section{Introduction}
\label{sec:introduction}

Lowering human greenhouse gases emissions is one major goal of the efforts against climate change, and the focus and concern of international cooperation (\href{https://unfccc.int/process-and-meetings/the-paris-agreement/the-paris-agreement}{Paris Agreement, 2015}). Many indicators of human development - population, Gross Domestic Product (GDP), environmental footprint - have been following exponential curves during the past decades \cite{acceleration}; hence, drastic measures are needed if we are to switch from increasing to quickly decreasing emissions, as expressed in global organisations goals (IPCC Fifth Assessment Report \yrcite{IPCC}).

Understanding and forecasting the evolution, on a country-scale, of various indicators related to carbon emissions, may help to give a clear idea of the progress we are making, or not, towards lower emissions. The main indicators that we chose to study are the variables appearing in Kaya identity \cite{kaya}, on a country level: population, national GDP, energy supply and CO2 emissions. % ; for each of these indicators, we have data on a country level on a time range spreading from 1971 to 2019 (\ref{sec:datasets}).

Our main objective is to develop a model able to use this data to make accurate forecasts, on a medium/long-time horizon. Machine-learning models offer interesting advantages in comparison of traditional methods used for this type of work, typically statistical models \cite{cerqueira}. In particular, the recent development of Neural Ordinary Differential Equations offers a promising perspective in this case \cite{NODEs} - we explain the reasons for this choice in \ref{motivations}. We adapted Neural ODEs for this problem and compared the performance with a baseline statistical model.

% A secondary objective relates to the modelling of specific scenarios regarding the evolution of the indicators. This is a common interesting idea when studying mitigation against climate change, and allows to make the connection between public policies, global effort, and actual effect on climate change \cite{SSPs, SDGs}. We studied how fictive observations, independent to the training set, could be incorporated into a Neural ODE model in order to simulate particular scenarios.

\section{Kaya identity \& impact on climate change}
\label{sec:kaya}

The Kaya identity expresses a very simple relation between carbon emissions $F$, energy supply $E$, Gross Domestic Product (GDP) $G$ and population $P$ : $\quad F = P \times \frac{G}{P} \times \frac{E}{G} \times \frac{F}{E}$.\\$\frac{G}{P}$ is the GDP per capita, representing our average life standard, which we generally want to increase; $\frac{E}{G}$ is the energy intensity of the GDP - the energy needed to create one unit of GDP -, it quantifies the efficiency of energy usage in our activities; $\frac{E}{F}$ is the carbon intensity of energy - the CO2 emission corresponding to the supply of one unit of primary energy - and indicates how reliant on carbon emissions our electricity production is.

Forecasting human development and carbon emissions is possible both with the set of raw variables $\{P, G, E, F\}$ and with the set of indicators appearing in the Kaya identity $\{P, \frac{G}{P}, \frac{E}{G}, \frac{F}{E}\}$. However, the latter gives a clearer analysis, from a macroscopic point of view \cite{kayaanalysis}. While the raw variables are very strongly correlated altogether and vary greatly between countries, the variables from the Kaya equation look more like consistent indicators actionable under the right choice of policies \cite{kayaindependance}. Overall, using these four indicators seems to be a good choice in order to assess efforts made by a country or region concerning carbon emissions \cite{kayabaltics}.

% Forecasting human development and carbon emissions is possible both with the set of raw variables $\{P, G, E, F\}$ and with the set of indicators appearing in Kaya identity $\{P, \frac{G}{P}, \frac{E}{G}, \frac{F}{E}\}$. However, the latter gives a clearer analysis, from a macroscopic point of view \cite{kayaanalysis}. While the raw variables are strongly correlated altogether, the indicators from Kaya equation are more likely to be independent variables, actionable under the right choice of policies : the GDP per capita represents our average life standard, which we generally want to increase ; the energy intensity of the GDP indicates if energy is used efficiently in our economic activities (saving energy versus wasting energy) ; finally, the carbon intensity represents how reliant on carbon emissions our electricity production is.

% In particular, Kaya identity shows clearly that, even if an economy improves its energy intensity and / or its carbon intensity, its emissions can still follow an increasing trend, for example if population growth or GDP growth is too high ; this is linked to the rebound effect, which can lead to misunderstandings regarding the effects of public policies \cite{rebound}. Overall, using these four indicators seems to be a good choice in order to assess efforts made by a country or region concerning carbon emissions \cite{kayabaltics}.

\section{Methodology}

\subsection{Datasets}
\label{sec:datasets}

% Four datasets - one for each raw variable - were used in this study ; indicators appearing in Kaya identity were computed from these raw variables afterwards. All datasets offer data for at least 54 countries, from 1971 to 2019. 

% Population \cite{WBPop} and GDP \cite{WBGDP} datasets were collected from the World Bank Open Data platform.

% The energy supply was extracted from a public dataset from the International Energy Agency \cite{IEABalances}. The retained value is the total energy supply, which corresponds to national production plus imports minus exports, summed for all studied products (kilotonnes of oil equivalent).

% The CO2 emission was extracted from another IEA public dataset \cite{IEAEmissions}. The retained value is the total CO2 emission from fuel combustion (million tonnes of CO2). It should be noted that this is not the total emission for each country, only the emission from fuel combustion - however, greenhouse gases emitted by fuel combustion represent around 75\% of all greenhouse gases emissions \cite{emissionsanalysis}.

Four datasets were used for this study: datasets for population \cite{WBPop} and GDP \cite{WBGDP} were collected from the World Bank Open Data platform; the total energy supply \cite{IEABalances} and the CO2 emissions from fuel combustion \cite{IEAEmissions} were extracted from public datasets from the International Energy Agency. It should be noted that this is not the total emission for each country - however, greenhouse gases emitted by fuel combustion represent around 75\% of all greenhouse gases emissions \cite{emissionsanalysis}. Each variable is available yearly, from 1971 to 2019, for at least 54 countries.

\subsection{Motivations for a machine-learning approach}
\label{motivations}

In the domain of energy and emissions, forecasts are often relying on expert knowledge and statistical models \cite{stirpat, statUS}; an extensive set of Integrated Assessment Models are used to explore how system Earth evolves during the next decades, following specific scenarios (IPCC Fifth Assessment Report \yrcite{IPCC}), \cite{AfricaIAM}. In other fields, grey models \cite{greymodel} have been used successfully to model future emissions, linking with population, GDP and energy supply \cite{greymodelChina}.

In some cases, black-box, data-driven models give good performances in comparison with traditional models \cite{MLvsstat}. Here, the problem naturally appears as a multivariate time-series forecasting problem, for which machine-learning already offers an extensive toolbox \cite{LSTMs, GRUs}.

Here, the variables that we are trying to predict are physical values, that may be independent, or follow a simple or complex relationship \cite{kayaindependance}. Since we may lack understanding of the physical system, a black-box model comes with the advantage of sparing the risk of making wrong hypotheses. In addition, even if the individual time series are only 50 years long, the dataset comprises a large set of countries to train a model with. Hopefully, a suitable machine-learning model would benefit from the variety of countries and produce more satisfying results training on several countries (\ref{multicountry}).

\subsection{Neural ODEs}

Neural Ordinary Differential Equations \cite{NODEs} are a type of models recently brought into the spotlight. The model learns the dynamics of a system with a neural network: if $X(t)$ is a state vector obeying dynamics of the shape $\frac{dX}{dt} = f(X, t)$, the network must approximate $f$. NODEs work together with a Differential Equation solver, and were presented with a practical method to compute a loss and back-propagate its gradient. Since the architecture of the neural network is entirely free, a NODE model can be used for very different problems.

In our case, NODEs offer a good extrapolation capacity, allowing us to make long-time forecasts. Since they naturally model a vector field representing the evolution of a physical system, we hope that such a model can capture the complex physical pattern existing between the forecast indicators. In addition, several countries could help the model to represent the dynamics more accurately. Finally, a NODE model can handle sporadic data (not regularly sampled), both in time and across dimensions \cite{GRUODE} - although not required with our original dataset, this property may become relevant for this problem (\ref{scenario}).

\section{Forecasting results}
\label{forecasting}

We used a NODE model to forecast the evolution of 7 variables, for each country. Apart from the 4 indicators from Kaya identity (population, GDP per capita, energy intensity of the GDP, carbon intensity of the energy), we added 3 variables representing how electricity is produced: proportion of electricity produced via fossil fuels combustion, via nuclear power or via renewable sources. The last 3 variables remain in $[0,1]$ at all time, and sum up to $1$.
Our model wraps a very simple neural network (one hidden layer), which takes $8$ inputs ($7$ variables plus time) and outputs $7$ values (the $7$ variables derivatives). We also tried deeper architectures ; without clear evidence of better performances on this simple case, we kept the simplest model for performance evaluation. We used an Adam optimisation strategy \cite{adam}.

\begin{figure}
\vskip -0.05in
\begin{center}
\centerline{\includegraphics[width=1\linewidth]{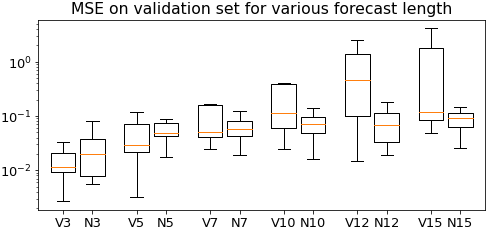}}
\caption{Boxplots of performances on 15 countries. \textit{V} stands for VAR model, \textit{N} stands for NODE model, the number corresponds to the forecast length, in years. We manually tuned some parameters. All models were trained for the same number of epochs, using Adam optimiser.}
\label{performances}
\end{center}
\vskip -0.5in
\end{figure}

\begin{figure*}
\vskip -0.05in
\begin{center}
\centerline{\includegraphics[width=1\linewidth]{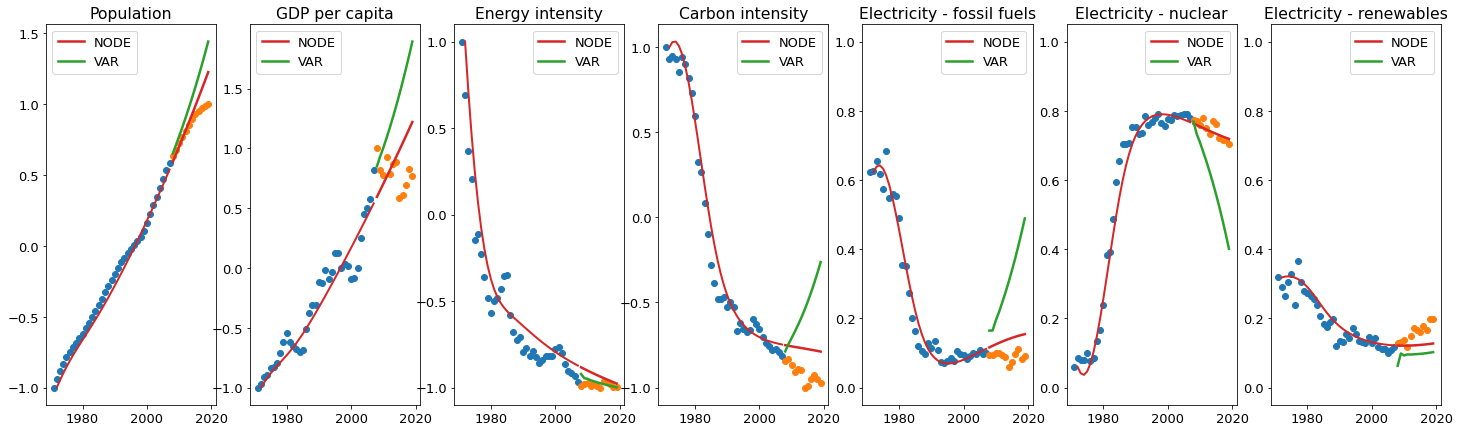}}
\caption{Example of forecasts obtained from a NODE model (red) versus a VAR model (green) ; training set is France data between 1971 and 2008, validation set is data between 2008 and 2020 (12-years forecast).}% In this case, the VAR model performed surprisingly bad ; the mean squared error on the validation set is ten times greater with VAR than with the NODE model.}
\label{forecasts}
\end{center}
\vskip -0.4in
\end{figure*}

As a first step, we ran the model for each country individually, dividing the rime range $[1971; 2019]$ in a training set (earlier dates) and validation set (latest dates). We compared the NODE model with a very simple statistical Vector Autoregression model (VAR model), which can typically be used for multivariate time-series problems \cite{VAR}.

To quantify performances, we computed the mean squared error on the validation set for a set of 15 European countries, for both models; figure \ref{performances} reports these results using boxplots. This was done for several forecast lengths. In particular, the table indicates that NODE models really shine for long-time horizons, while for very short forecasts VAR performs better; NODE models also give more stable performances - lower variance - when the forecast length increases. In addition, figure \ref{forecasts} presents a typical output of our NODE model, compared with forecasts from a VAR model. 

\section{Conclusion \& next steps}

According to our results, Neural ODEs give overall good performances compared with VAR models, especially for long-time horizons. This justifies our approach as a reasonable method to forecast emissions in the long term. Through a data-driven model only, without expert knowledge, this process would hopefully learn from present and past patterns in the evolution of indicators from the Kaya identity, in order to give more accurate insight to policymakers.

\subsection{Multi-country training}
\label{multicountry}

One of the motivations for a machine-learning solution to this problem is the perspective of benefiting from training the same model on several countries data. We started from the assumption that countries with similar profiles - e.g. same continent, same economic system, etc. - should have their forecast indicators obey the same hidden law; in other words, the system dynamics should be identical for these countries \cite{correlationsmulticountry, economycorrelations}. If this hypothesis is true, crossing data from these countries allows to train the model on more samples0 and, hopefully, the trained model will be more robust and able to forecast a broader set of behaviours. %A simple illustration would be : if the model learns how indicators were correlated after an economic crisis from country A, it should be able to make more accurate forecasts for country B for which an economic crisis is just starting.

In the present state, we need to rethink how our model is able to distinguish different countries. With the normalisation process used so far, it was sometimes impossible for the model to tell that time-series come from different sources, with the output being just an average of all the observed trajectories. One possibility for this type of issue is to add meta-information about the data source, as additional inputs to the neural network \cite{hydro}. In our case, as first idea, we added country encoding as an input (one-hot encoding). This led to better forecasts, but work remains to be done in this direction ; we made preliminary experiments on deeper networks, which might also help.

\subsection{Scenario modelling}
\label{scenario}

A major prospect and objective of our study is the modelling of particular scenarios with this machine-learning approach: if we believe that our model has captured correctly the interactions between the forecast indicators, it is desirable to be able to explore particular possible futures \cite{RCPs, SSPs}. This is a common idea when studying mitigation against climate change, and allows to make the connection between public policies, global effort, and actual effect on climate change \cite{SDGs}. Here, interesting scenarios could model how emissions forecast change if the electricity production shifts towards more nuclear or renewable energy; or study trajectories that allow to meet national or international goals on emissions reductions; or model the impact of a crisis such as Covid-19 on long-term energy supply and emissions.

We envisioned two ways to incorporate scenarios to our NODE model - in both cases, we train the model beforehand. First, we could provide the model with a full, hypothetical trajectory for one indicator - the chosen scenario -, and forecast all other indicators. Hopefully, the model would take advantage of its training experience to output relevant correlations. Second, we could add after training one or several "hypothetical observations" to the training set, for future dates - the chosen scenario - ; we would then train the model for a few more epochs on the augmented training set, expecting the model to adapt its forecasts to the new observations, while keeping the structure it has acquired during the first training. A simple way to verify this idea would be to provide the model with observations from the validation set, and to examine if the performance on the validation set increases significantly after the additional training. Further work is required to assess if this approach can lead to consistent results.

\bibliography{main}
\bibliographystyle{icml2021}

\end{document}